\documentclass[letterpaper]{article}
\usepackage{aaai18}
\usepackage{times}
\usepackage{helvet}
\usepackage{courier}
\usepackage{url}
\usepackage{siunitx}
\usepackage{graphicx}
\usepackage{subcaption}
\usepackage{amsmath}
\usepackage{amssymb}
\usepackage{amsthm}
\usepackage{algorithm}
\usepackage{algpseudocode}
\usepackage{optidef}
\usepackage{mathtools}
\usepackage{multirow}
\usepackage{array}
\usepackage{diagbox}
\usepackage{breqn}
\newcommand{\citet}[1]
{\citeauthor{#1}~\shortcite{#1}}
\newcommand{\citep}{\cite}

\title{Better Safe than Sorry: Evidence Accumulation Allows\\ for Safe Reinforcement Learning}
\author{Akshat Agarwal\thanks{Denotes equal contribution}
\thanks{Corresponding Author: \texttt{aa7@cmu.edu}}\\
Robotics Institute\\
Carnegie Mellon University\\
\And Abhinau Kumar V\footnotemark[1]\\
Department of Electrical Engineering\\
Indian Institute of Technology Hyderabad\\
\AND Kyle Dunovan, Erik Peterson, Timothy Verstynen\\
Department of Psychology\\
Carnegie Mellon University\\
\And Katia Sycara\\
Robotics Institute\\
Carnegie Mellon University\\
}

\begin{document}
\maketitle

\begin{abstract}
In the real world, agents often have to operate in situations with incomplete information, limited sensing capabilities, and inherently stochastic environments, making individual observations incomplete and unreliable. Moreover, in many situations it is preferable to delay a decision rather than run the risk of making a bad decision. In such situations it is necessary to aggregate information before taking an action; however, most state of the art reinforcement learning (RL) algorithms are biased towards taking actions \textit{at every time step}, even if the agent is not particularly confident in its chosen action. This lack of caution can lead the agent to make critical mistakes, regardless of prior experience and acclimation to the environment. Motivated by theories of dynamic resolution of uncertainty during decision making in biological brains, we propose a simple accumulator module which accumulates evidence in favor of each possible decision, encodes uncertainty as a dynamic competition between actions, and acts on the environment only when it is sufficiently confident in the chosen action. The agent makes no decision by default, and the burden of proof to make a decision falls on the policy to accrue evidence strongly in favor of a single decision. Our results show that this accumulator module achieves near-optimal performance on a simple guessing game, far outperforming deep recurrent networks using traditional, forced action selection policies.

\end{abstract}
% Todo: EvAc? (as a short-hand name)

\section{Introduction}
% Introduce the problem setting
Traditional reinforcement learning (RL) algorithms map the state of the world to an action so as to maximize a reward signal it receives from environmental feedback. With the success of deep RL, this action-value mapping is increasingly being approximated by deep neural networks to maximize success in complex tasks, such as Atari games \cite{mnih2015human} and Go \cite{silver2016mastering}.
In the real world, RL agents usually operate with incomplete information about their surroundings due to a range of issues issues such as limited sensor coverage, noisy data, occlusions, and the inherent randomness in the environment which comes from factors that can't be modeled. With individual observations being incomplete and/or unreliable, it is imperative that agents accrue sufficient evidence in order to make the most task or environmentally appropriate decision. In current RL approaches, this is accomplished by using recurrent layers in the neural network to aggregate information \cite{lample2017playing,agarwal2018community,zoph2016neural}; however, this pipeline is biased towards taking a decision at every time step, even if the agent is not particularly confident in any of the possible actions. This is highly undesirable, especially in situations where an incorrect action could be catastrophic (or very heavily penalized). The usual mechanism to allow the possibility of not taking an action at any time step is to have one possible action as a `No-Op', which can be chosen by the agent when it does not want to act on the environment. This requires the policy to actively choose to not act, which is a counter-intuitive requirement for any real world scenario, where not taking an action should be the default setting.

In biological networks, the circuit-level computations of Cortico-Basal-Ganglia-Thalamic (CBGT) pathways \cite{mink1996basal} are ideally suited for performing the multiple sequential probability ratio test (MSPRT) \citep{wald45sprt,bogacz2007basal,bogacz2007optimal}, a simple algorithm of information integration that optimally selects single actions from a competing set of alternatives based on differences in input evidence \citep{draglia1999multihypothesis,baum1994sequential}. 
% Allowing dopaminergic learning signals to implement a form of RL by modifying the efficacy of cortical inputs into the CBGT pathways produces an adaptive variant of the MSPRT that approximates the optimal solution to the action selection process based on both sensory signals and feedback learning \citep{bogacz2011integration,Caballero2018-vs}. 
Motivated by these theories of dynamic resolution of decision uncertainty in the CBGT pathways in mammalian brains (see also \citep{redgrave1999basal,mink1996basal,dunovan2016believer}), \textit{we propose modifying existing RL architectures by replacing the policy/Q-value output layers with an accumulator module} that makes a decision by accumulating the evidence for alternative actions until a threshold is met. Each possible action is represented by a channel through which environmental input is sampled and accumulated as evidence at each time step, and an action is chosen only when the evidence in one of the channels crosses a certain threshold. This ensures that when the environment is stochastic and uncertainty is high, the agent can exercise greater caution by postponing the decision to act until sufficient evidence has been accumulated, thereby avoiding catastrophic outcomes. While evidence accumulation necessarily comes at a cost to decision speed, there are many real world scenarios in which longer decision times are considered a perfectly acceptable price to pay for assurances that those decisions will be both safe and accurate. 

The accumulator module can work with both tabular and deep reinforcement learning, with on-policy and off-policy RL algorithms, and can be trained via backpropagation. We present a simple guessing task where the environment is partially observable, and show that a state of the art RL algorithm (A2C-RNN \cite{mnih2016asynchronous}) fails to learn the task when using traditional, forced action selection policies (even when equipped with a `No-Op'), but achieves near-optimal performance when allowed to accumulate evidence before acting.

\section{Related Work}
% drqn etc, HMMs?
Partially Observable Markov Decision Processes (POMDPs) \cite{kaelbling1998planning,jaakkola1995reinforcement,kimura1997reinforcement} are the de-facto choice for modeling partially observable stochastic domains. \citet{hausknecht2015deep} first successfully used an LSTM layer in a DQN \cite{mnih2015human}. Since then, it has become a standard part of Deep RL architectures, including both on-policy and off-policy RL algorithms \cite{agarwal2018challenges,lample2017playing}. Another strategy consists of using Hidden Markov Models \cite{monahan1982state} to learn a model of the environment, for domains where the environment is itself Markovian, but does not appear to be so to the agent because of partial observability.

% spiking networks and zambrano
The implementation of accumulation-to-threshold dynamics in single neurons, where inputs are accumulated over time as a sub-threshold change in potential until a threshold is reached, causing the cell to "fire", has been studied to a great extent as spiking neural networks \cite{o2016deep} for supervised learning using backpropagation \cite{lee2016training}. Each neuron in the neural network is replaced by a Stochastic/Leaky Integrate-and-Fire neuron, with Winner-Take-All (WTA) circuits. \citet{florian2007reinforcement} presented a reinforcement learning algorithm for spiking networks through modulation of spike timing-dependent plasticity. \citet{zambrano2015continuous} also presented a continuous-time on-policy RL algorithm to learn task-specific working memory in order to decouple action duration from the internal time-steps of the RL model using a Winner-Take-All action selection mechanism. The approach taken here differs from these previous examples in three key ways: first, we consider simple additive accumulators without any leakage; second, the dynamic competition between neurons is modeled using center-surround inhibition, allowing between-channel dynamics to modulate the evidence criterion; and third, evidence accumulation is restricted to neurons in the last (e.g., output) layer of the network. 

% lipton catastrophic states, related safety work
Prior safe reinforcement learning models \cite{garcia2015comprehensive} have primarily been divided into two lines of work - the first is based on modification of optimality criteria to incorporate worst-case criteria, or risk-sensitive criteria. The second line focuses on the modification of the exploration process through incorporation of external knowledge, teacher guidance or risk-directed exploration. Recently, \citet{lipton2016combating} used intrinsic fear and reward shaping to learn and avoid a set of dangerous (catastrophic) states, in the context of lifelong RL. \citet{chow2018lyapunov} used Lyapunov functions to guarantee the safety of a behavior policy during training via a set of local, linear constraints. These works focus on a different aspect of safety in reinforcement learning, and are complementary to ours.

\begin{figure}
    \centering
    \includegraphics[width=\columnwidth]{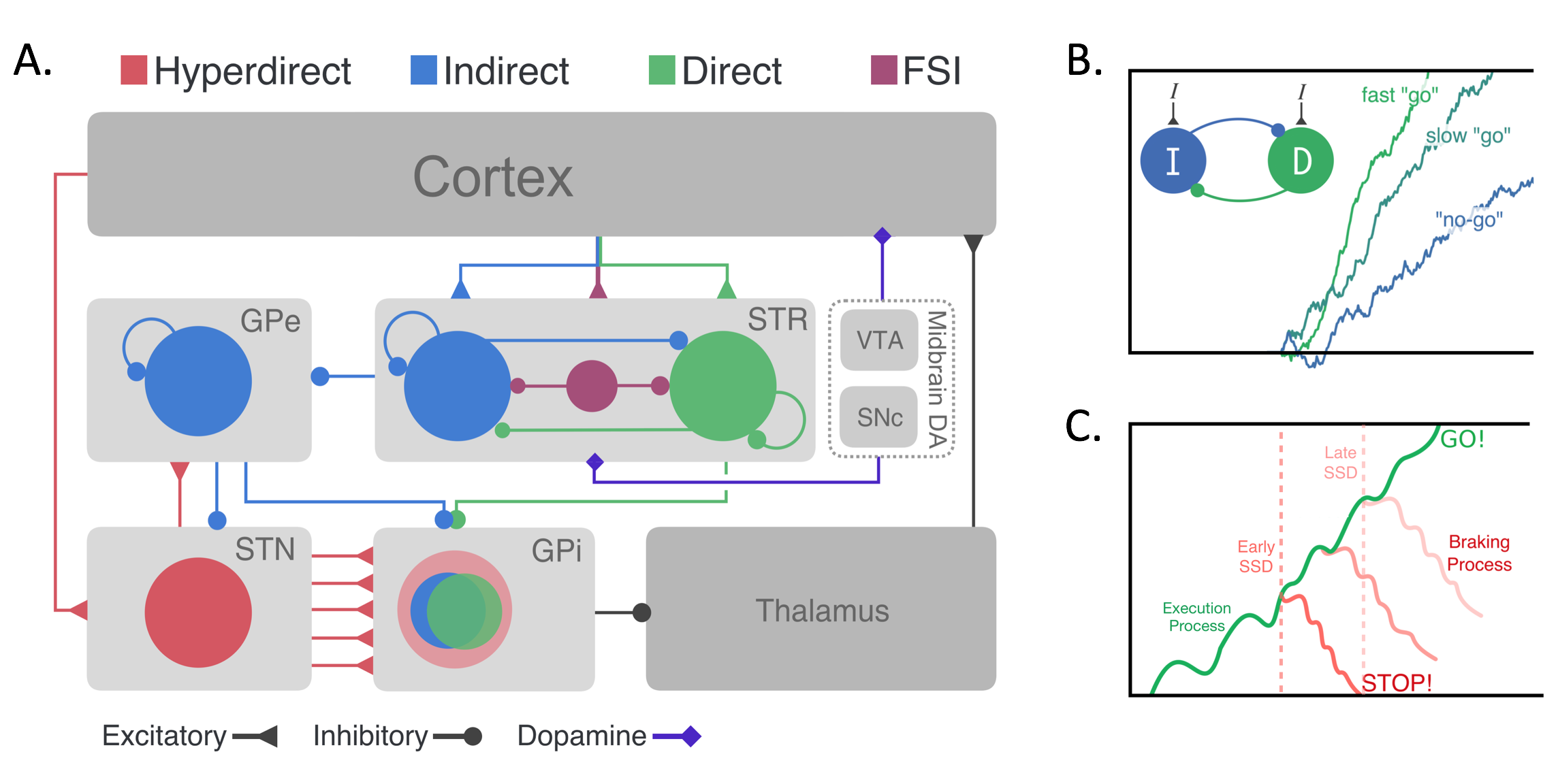}
    \caption{\textbf{Cortico-Basal-Ganglia-Thalamus (CBGT) networks \& the dependent process accumulator model.} (A) CBGT circuit. Striatum, STR; fast-spiking interneurons, FSI; globus pallidus external, GPE; globus pallidus internal (GPi); subthalamic nucleus, STN; ventral tegmental area, VTA; substantia nigra pars compacta, SNc. (B) Competing pathways model of the direct (D) \& indirect (I) pathways. (C) The dependent process model. Panels A and B were recreated with permission from \cite{dunovan2016believer} and Panel C was recreated with permission from \cite{dunovan2015competing}.}
    \label{fig:cbgt}
\end{figure}

\section{Decision Making in the Brain}
\label{sec:cbgt}
We now briefly describe the dependent process model of decision making in the brain, which serves as the biological inspiration for evidence accumulation in reinforcement learning. 
Decision making in CBGT circuits can be modeled as an interaction of three parallel pathways: the direct pathway, the indirect pathway and the hyperdirect pathway (see Fig.\ref{fig:cbgt}A). The direct and indirect pathways act as gates for action selection, with direct pathway facilitating and indirect pathway inhibiting action selection. These pathways converge on a common output nucleus (the GPi). From a computational perspective, this convergence of the direct and indirect pathways suggests that their competition encodes the rate of evidence accumulation in favor of a given action \cite{bogacz2007basal,dunovan2016believer}, resulting in action execution if the direct pathway sufficiently overpowers the indirect pathway. The decision speed is thus modulated by the degree of response conflict across actions (see Fig. \ref{fig:cbgt}B). A second inhibitory pathway (the hyperdirect pathway) globally suppresses all action decisions when the system is going to make an inappropriate response, with the competition between all three major pathways formalized by the so-called dependent process model of CBGT computations (see Fig. \ref{fig:cbgt}C) \cite{dunovan2015competing}. The center-surround architecture of the CBGT network (such that inputs to a direct pathway for one action also excite indirect pathways for alternative actions) \citep{mink1996basal}, as well as the competitive nature of direct and indirect pathways within a single action channel \citep{bogacz2007basal,dunovan2016believer} allows for modulation of both the rate and threshold of the evidence accumulation process \cite{dunovan2017errors}. Moreover, this selection mechanism implicitly handles situations in which no action is required from the agent, as evidence simply remains at sub-threshold levels until a significant change is registered in the environment \cite{dunovan2016believer}. 

This dynamic selection process runs in sharp contrast to standard Deep RL methods \cite{mnih2015human,mnih2016asynchronous}. Deep RL operates at a fixed frame rate, mandating actions to be taken with a particular fixed frequency even in times of high uncertainty or times when actions might not be needed. Deep RL uses backpropagation to modulate representations of state-value and action-value (a process analogous, but not identical, to how the dopamine projections to cortex alter cortical representations), whereas the actual gating units are static units with no feedback-dependent plasticity. We posit that incorporating additional plasticity at the output layer holds significant promise to improving existing Deep RL algorithms, as adaption of the selection process will interact with the action representation process to facilitate complex action repertoires. 

\section{Methods}

\subsection{Mode Estimation (ME) Task}
\begin{figure}
    \centering
    \includegraphics[width=\columnwidth]{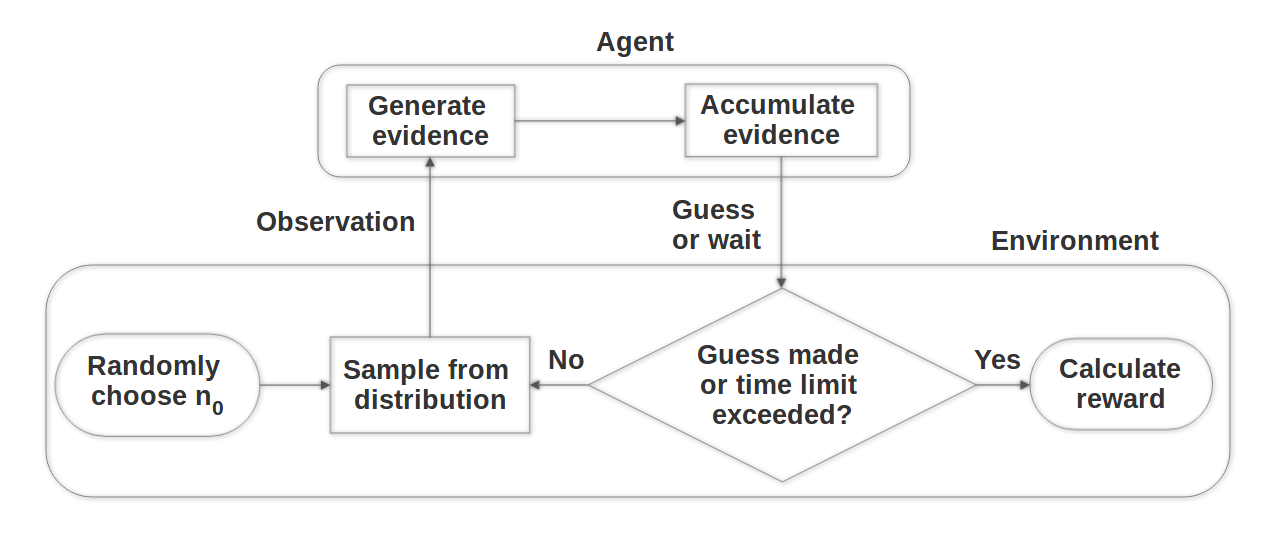}
    \caption{Flowchart describing agent-environment interaction}
    \label{fig:taskflow}
\end{figure}

We propose a simple episodic task where the agent receives a sample from a discrete unimodal distribution at each time step, and has to estimate the mode of the distribution. Within each episode, the agent receives a sample from the environment distribution at each step, and has the choice to make a decision (estimate the mode) or not. If the agent chooses not to make a decision at that time step, it simply receives another sample from the environment. The episode ends either when the agent makes a guess, or if the maximum allowed length of an episode, $T_{max}$, is exceeded. The beginning of each new episode resets the environment and randomly changes the distribution from
which the samples being observed by the agent are generated. 

Conditioned on the spread of the hidden environmental distribution, the agent must learn to delay its decision and aggregate samples over multiple time steps to make an informed estimate of the mode. The agent receives feedback from the environment in the form of a reward $r_{t}$ at the end of each episode, such that the agent is rewarded for making the correct decision and penalized for tardiness and for making an incorrect decision (or no decision at all).

\begin{equation}
r_{t} = \begin{cases}
0 & \text{$t \leq T_{max}$ \textbf{and} No guess}\\
R_{1} - (t - 1) & \text{$t \leq T_{max}$ \textbf{and} Correct guess}\\
R_{2} & \text{$t \leq T_{max}$ \textbf{and} Incorrect guess} \\
R_{3} & \text{$t > T_{max}$} 
\end{cases}
\label{eq:reward}
\end{equation}
where $t$ is the number of time steps the agent waited and accumulated information before making a guess. $R_{1}$ is the reward the agent receives if it guesses correctly at the first time step, $R_{2}$ is the penalty the agent receives for making an incorrect guess and $R_{3}$ for not making a guess at all within $T_{max}$ time steps. The reward received by the agent decays linearly with the number of time steps it waits before making a decision, requiring it to balance the trade-off between making a decision quickly and making a decision accurately. This becomes especially relevant at higher values of $\epsilon$, when the observations are very noisy. The interaction between the agent and the environment is illustrated in Fig.\ref{fig:taskflow}.

\subsection{The Accumulator Module}
At each time step, the agent receives an observation $o_t$ from the environment, from which it extracts an evidence vector $\kappa_t = f(o_t), \kappa_t \in \mathbb{R}^{n}$, consisting of one component for each available action.
The cumulative evidence received since the beginning of each episode ($t = 0$) is stored in accumulator channels $\nu$
\begin{equation}
    \nu^i = \sum_{t=0}\kappa^i_t
\end{equation}
where $\kappa^i_t$ is the i-th component of the evidence vector $\kappa_t$.
The preference vector $\rho$ over the action choices is given by a softmax over $\nu$, such that 
\begin{equation}
    \rho^{i} = \dfrac{\exp({\nu^{i}})}{\sum_{i=1}^{A}\exp({\nu^{i}})}
\end{equation}
$\rho$ encodes the agent's confidence in actions - if the accumulated evidence values $\nu_{i}$ are high for multiple actions, then the preference values $\rho_{i}$ for all of them will be relatively low, indicating that the agent is not very confident in any particular action choice. Since we do not want any decision to be made in such a situation, action $i$ is taken only when $\rho_{i}$ crosses some threshold $\tau$, failing which no guess is made and the agent keeps observing more information from the environment, until the time when it becomes sufficiently confident to act on the environment. Note that the evidence accumulation process, as defined here, mirrors the hypothesis proposed in \citet{bogacz2007basal} regarding how the basal ganglia and cortex implement optimal decision making between alternative actions.

\subsection{Learning Algorithm}
We use the Advantage Actor-Critic (A2C) algorithm \cite{mnih2016asynchronous} to learn the RNN, the accumulator threshold $\tau$ and the evidence mapping $f$ in our experiments below. This is a policy gradient method, which performs an approximate gradient descent on the agent's discounted return $G_{t} = \sum_{k=0}^{\infty}\gamma^kr_{t+k+1}$. The A2C gradient is as follows:
\begin{dmath}
    (G_t-v_{\theta}(s_t))\nabla_{\theta}\log\pi_{\theta}(a_t|s_t) + \eta(G_t-v_{\theta}(s_t))\nabla_{\theta}v_{\theta}(s_t) + \beta\sum_{a}\pi_{\theta}(a|s)\log\pi_{\theta}(a|s)
\end{dmath}
where $s_{t}$ is the observation, $a_{t}$ the action selected by the policy $\pi_{\theta}$ defined by a deep neural network with parameters $\theta$ and $v_{\theta}(s)$ is a value function estimate of the expected return $\mathbb{E}[G_{t}|s_{t}=s]$ produced by the same network. Instead of the full return, we use a 1-step return $G_{t} = r_{t+1} + \gamma v_{\theta}(s_{t+1})$ in the gradient above. The last term regularizes the policy towards larger entropy, which promotes exploration, and $\beta$ is a hyper-parameter which controls the importance of entropy in the overall gradient. We keep the value loss coefficient $\eta$ and discount factor $\gamma$ fixed at 1 and 0.95, respectively, for all the experiments.

\section{Experiments and Results}

\begin{table*}
    \centering
    \begin{tabular}{|c|ccccc|}
        \hline
        \backslashbox{Agent}{Uncertainty $\epsilon$} & 0 & 0.2 & 0.4 & 0.6 & 0.8 \\
        \hline
        \textbf{Monte-Carlo Estimate} & 30 & 27.6 & 25 & 18.4 & -8.2 \\
        \textbf{A2C-RNN} & 30 & 26.9 & 7.5 & -23 & -30 \\
        \textbf{Learning $\tau$} & 30 & 27.6 & 24.9 & 17.4 & -13.7 \\
        \textbf{Joint Training of $\tau$ and $f$} & 29.7 & 25.7 & 22.2 & 12.5 & -25.2\\
        \hline
    \end{tabular}
    \caption{The expected rewards received by learning agents in environments with varying levels of uncertainty, after 50k episodes of training. The Monte-Carlo estimates provide a near-optimal baseline to compare the learning approaches with. It can be seen that the joint training method far outperforms A2C-RNN, while the agent learning only the accumulator threshold $\tau$ reaches near-optimal values close to the MC estimate.}
    \label{tab:reward}
\end{table*}

The simple mathematical structure of the Mode Estimation task allows us to find the optimal values of the accumulator threshold $\tau$ using Monte Carlo simulations, providing a good reference to compare the performance of our learning algorithms with. We first measure the performance of recurrent actor-critic policy gradient RL with a forced action-selection policy, verify that it is unable to learn anything meaningful when the environment stochasticity $\epsilon$ is high, and then demonstrate how using the accumulator module achieves near-optimal performance on a wide range of $\epsilon$ values. We train the accumulator threshold directly (with observations as evidence) with Advantage Actor-Critic (A2C) \cite{mnih2016asynchronous}, and then successfully jointly train deep networks to learn both the evidence mapping $f$ and accumulator threshold $\tau$ values using A2C.

\subsection{Task Instantiation}
In the particular instance of the Mode Estimation task we use for running experiments, the environment chooses an integer, say $n_{0}$, uniformly at random from the set of integers $\mathbb{Z} = \{0,1,2,\cdots,9\} $. Then, at each step during that episode, the agent receives an observation $n \in \mathbb{Z}$, with probability $p$ such that
\begin{equation}
p(n) = 
\begin{cases}
1 - \epsilon & \text{if n = $n_0$} \\
\dfrac{\epsilon}{9} & \text{otherwise}
\end{cases}
\label{eq:argmax}
\end{equation}
where $\epsilon$ is an environment parameter encoding the amount of randomness/noise inherent in the environment. The agent's task is to correctly guess the mode $n_{0}$ for that particular episode, based on these noisy observations. As soon as the agent makes a guess, the episode resets (a new $n_{0}$ is chosen). The reward received by the agent follows Eqn. \ref{eq:reward}, with $R_1 = -R_2 = -R_3 = 30$ and $T_{max} = 30$.

\subsection{Baseline Monte-Carlo Estimates of Accumulator Performance}
The advantage of using a simple task for evaluation is that we can obtain Monte-Carlo (MC) estimates of the best expected performance of the accumulator model for various values of the environment's randomness parameters $\epsilon$. The accumulator is parameterized by only one hyperparameter - the threshold $\tau $. Since $\tau $ is compared with components of the preference vector $\rho$, which are the output of a softmax operation, $\tau \in (0,1)$ covers the entire range of values which are useful. 
The agent receives observations in the form of one-hot vector representations of the distribution samples, which are directly treated as evidence to be accumulated. For each value of $\tau \in \{0,0.1,0.2,\ldots 0.9\}$, we complete 10,000 episode rollouts, tracking the reward received in each episode. The threshold with the highest expected reward is selected, and that reward is used as a near-optimal estimate of the accumulator's performance. This process is repeated for environments with varying levels of stochasticity, specifically, with $\epsilon \in \{0,0.2,0.4,0.6,0.8\}$.
In Fig. \ref{fig:allresults}, the accuracy, decision time and reward achieved by the optimal thresholds for each $\epsilon$ value are plotted with a dashed red line, and the expected reward received with the optimal thresholds is specified in Row 1 of Table \ref{tab:reward}.
Note that since we use the same discretization of $\tau$ (or a subset of it) when learning the threshold in subsequent sections, these are the highest possible rewards that our learning agents could receive, which is also reflected in Fig. \ref{fig:allresults}, where the rewards achieved by any learning agent never exceed the MC estimates.

\subsection{Recurrent A2C with Forced Action Selection}
Using traditional, forced action selection policies, we now train a recurrent policy for the Mode Estimation task. To allow a fair comparison with the evidence accumulators, the agent is given an additional `No-Op' action output which allows the agent the possibility of choosing not to make a decision and wait for more samples. We call this agent the `A2C-RNN' agent.
Note that recurrent policies are the state of the art method of dealing with partial observability in deep reinforcement learning \cite{hausknecht2015deep,mnih2016asynchronous}.

The policy network takes as input observations from the environment, and outputs (a) a probability distribution over the actions, and (b) a value function estimate of the expected return. The observation is a 4-dimensional binary representation of the sample (e.g. 0110 for 6), and is passed through a linear layer with a ReLU non-linearity to get an output of size 25. This goes through an RNN cell \cite{elman1990finding} of output size 25 with a ReLU non-linearity, and is then passed as input to two linear layers that output the probability distribution over actions (using a softmax activation) and the value estimate of the expected return. The agent has 11 possible action choices (including a `No-Op', and 10 choices for each of the possible modes). 

\begin{figure*}
        \centering
        \begin{subfigure}{0.6\textwidth}
                \includegraphics[width=\textwidth]{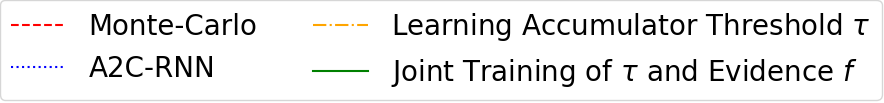}
                \phantomsubcaption
        \end{subfigure}
        
        \begin{subfigure}{0.32\textwidth}
                \includegraphics[width=\textwidth]{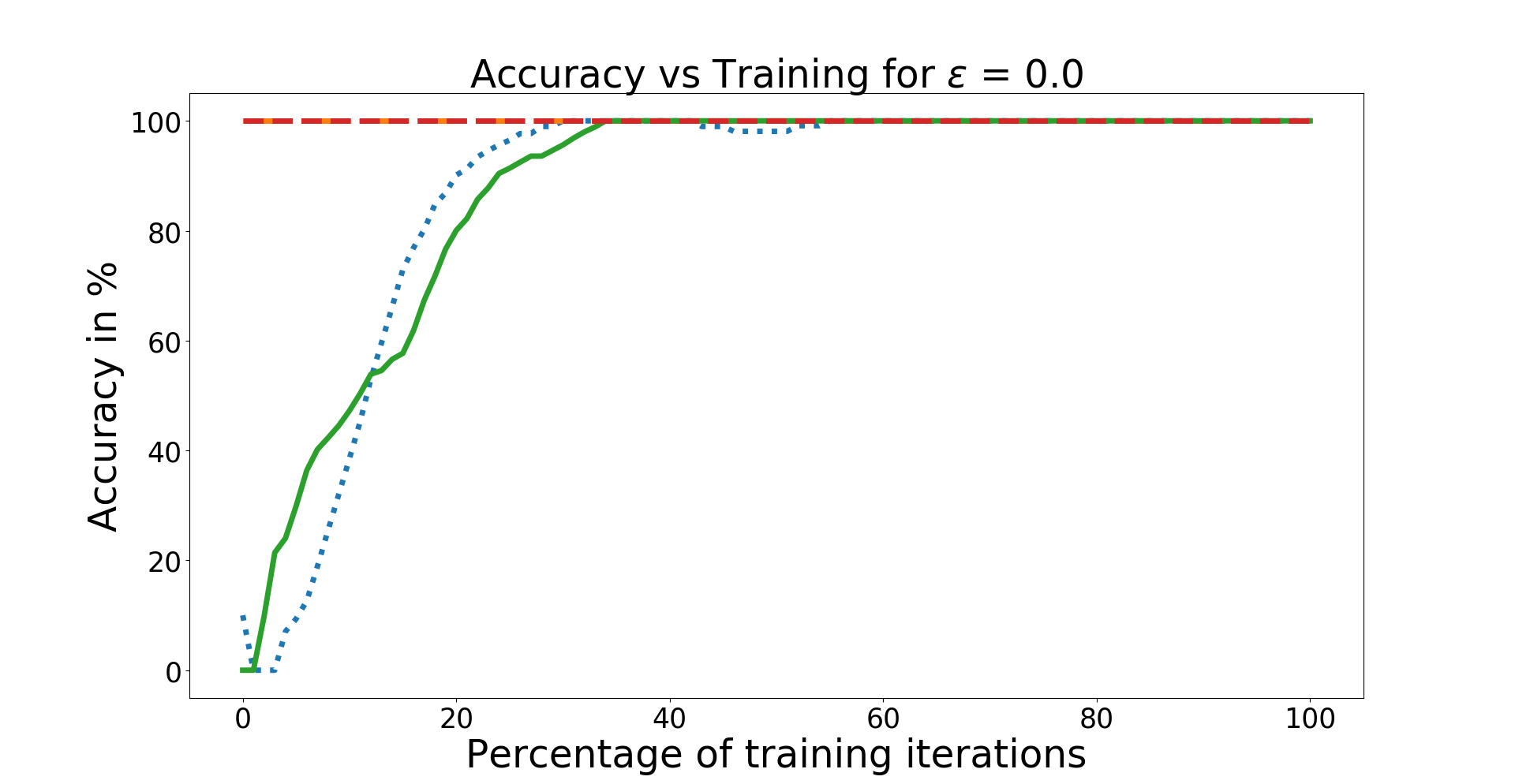}
                \phantomsubcaption
        \end{subfigure}
        ~
        \begin{subfigure}{0.32\textwidth}
                \includegraphics[width=\textwidth]{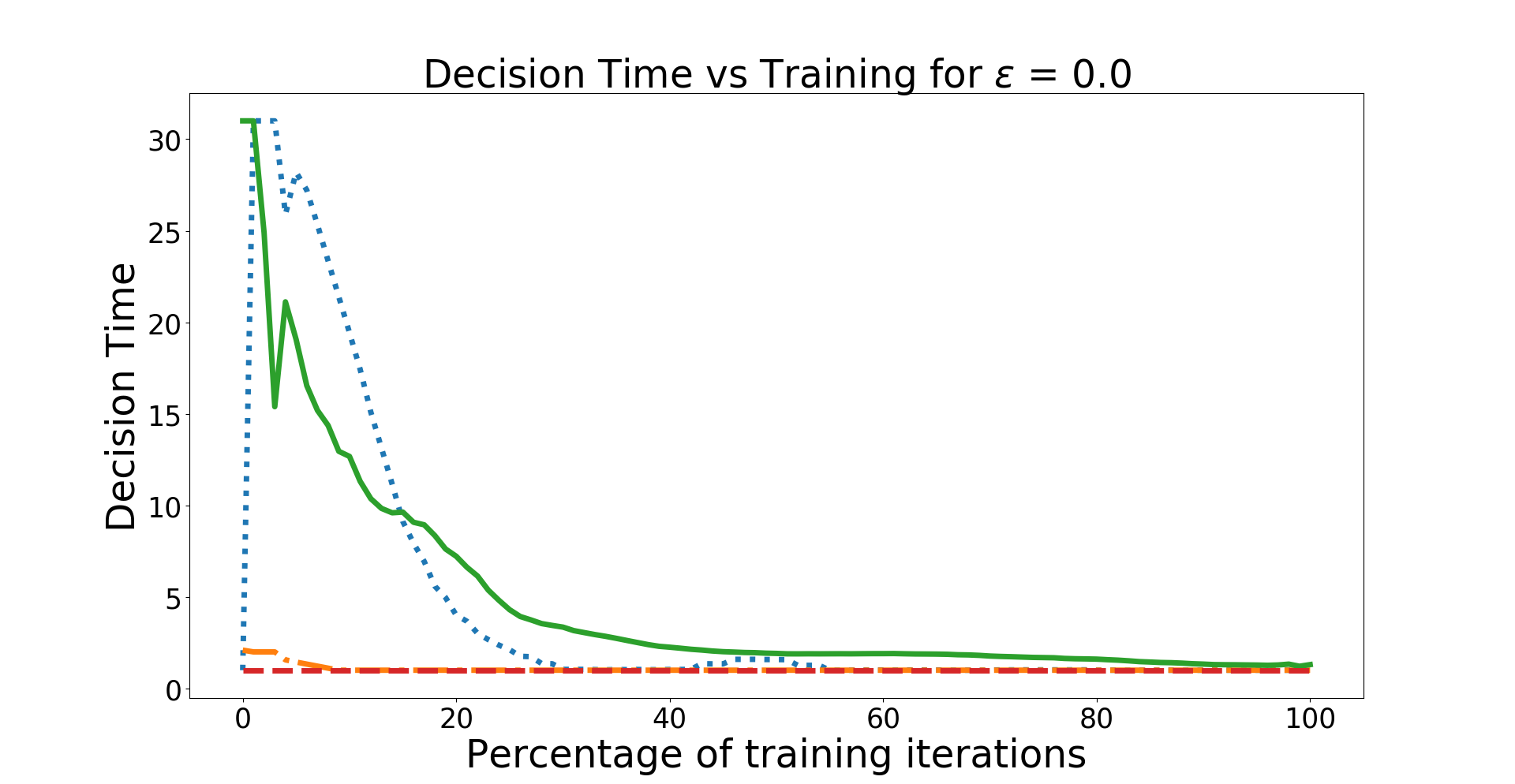}
                \phantomsubcaption
        \end{subfigure}
        ~
        \begin{subfigure}{0.32\textwidth}
                \includegraphics[width=\textwidth]{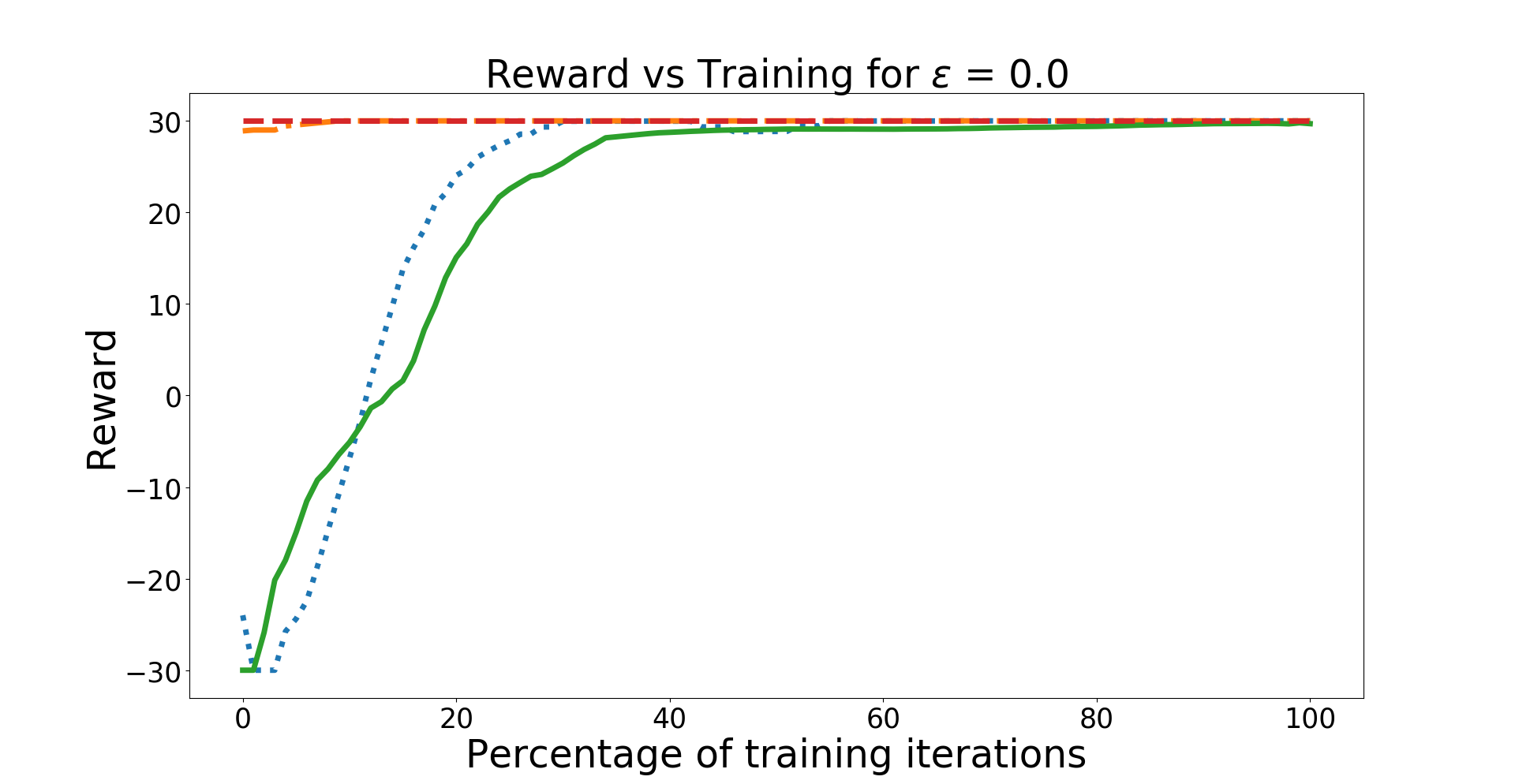}
                \phantomsubcaption
        \end{subfigure}
        
        \begin{subfigure}{0.32\textwidth}
                \includegraphics[width=\textwidth]{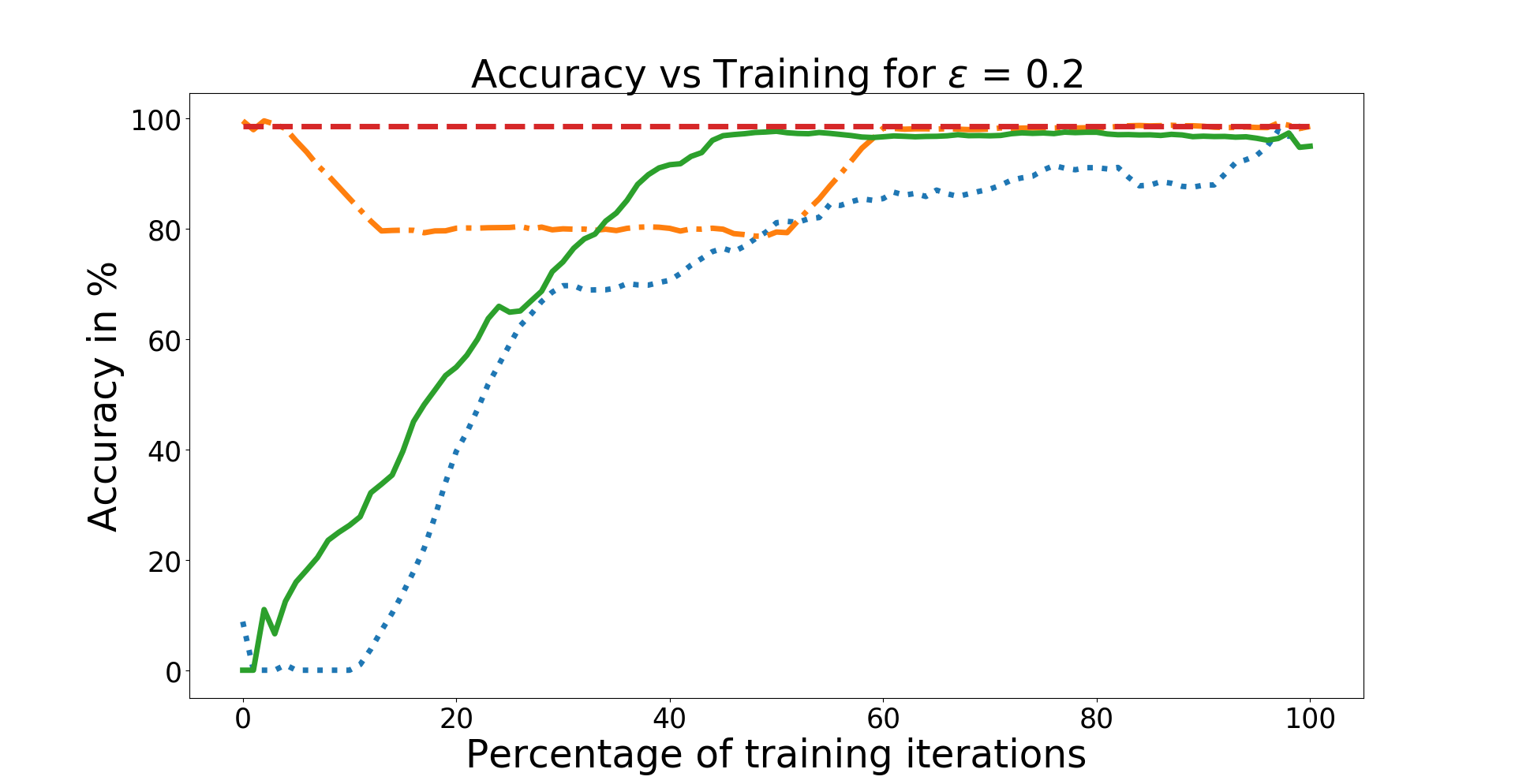}
                \phantomsubcaption
        \end{subfigure}
        ~
        \begin{subfigure}{0.32\textwidth}
                \includegraphics[width=\textwidth]{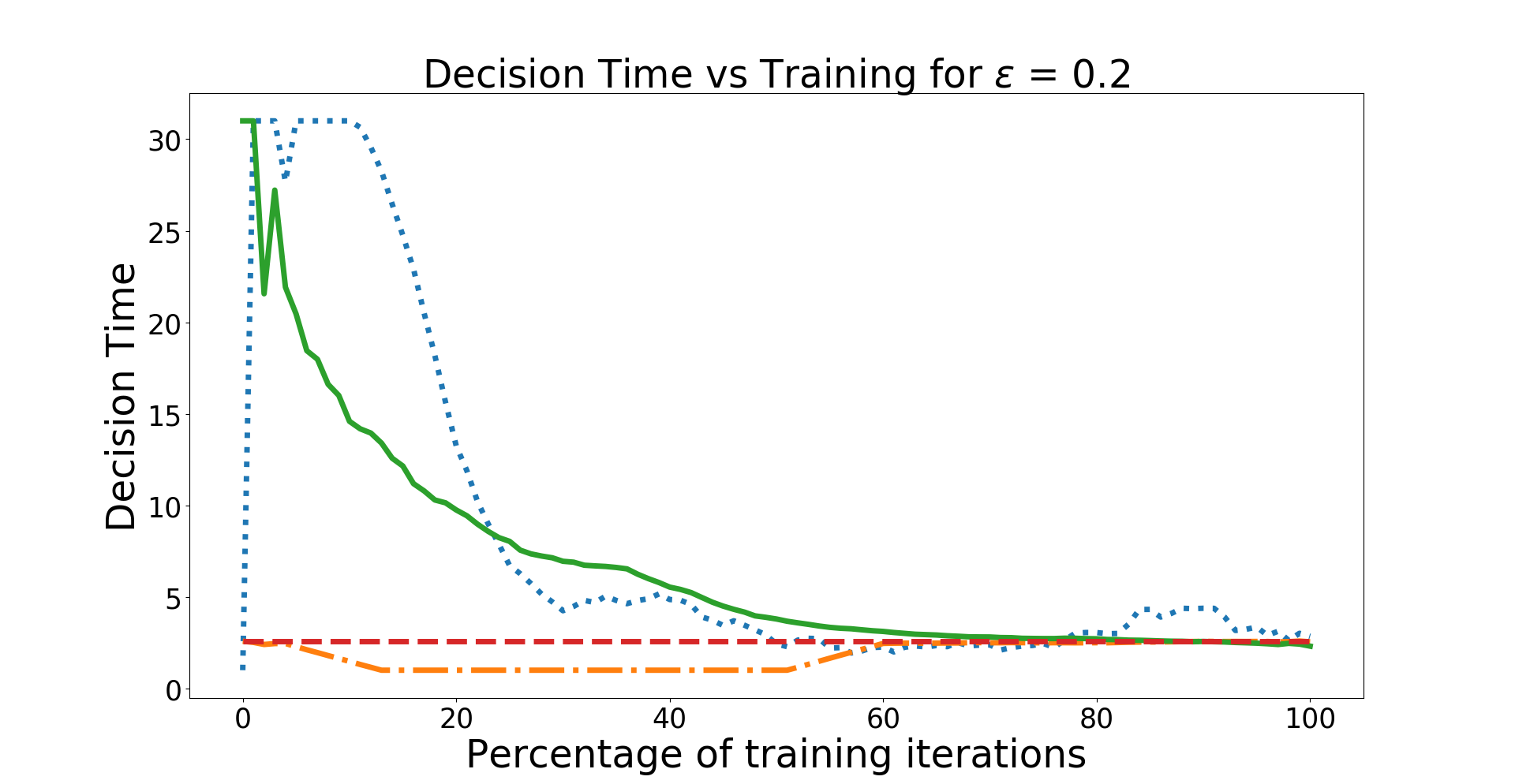}
                \phantomsubcaption
        \end{subfigure}
        ~
        \begin{subfigure}{0.32\textwidth}
                \includegraphics[width=\textwidth]{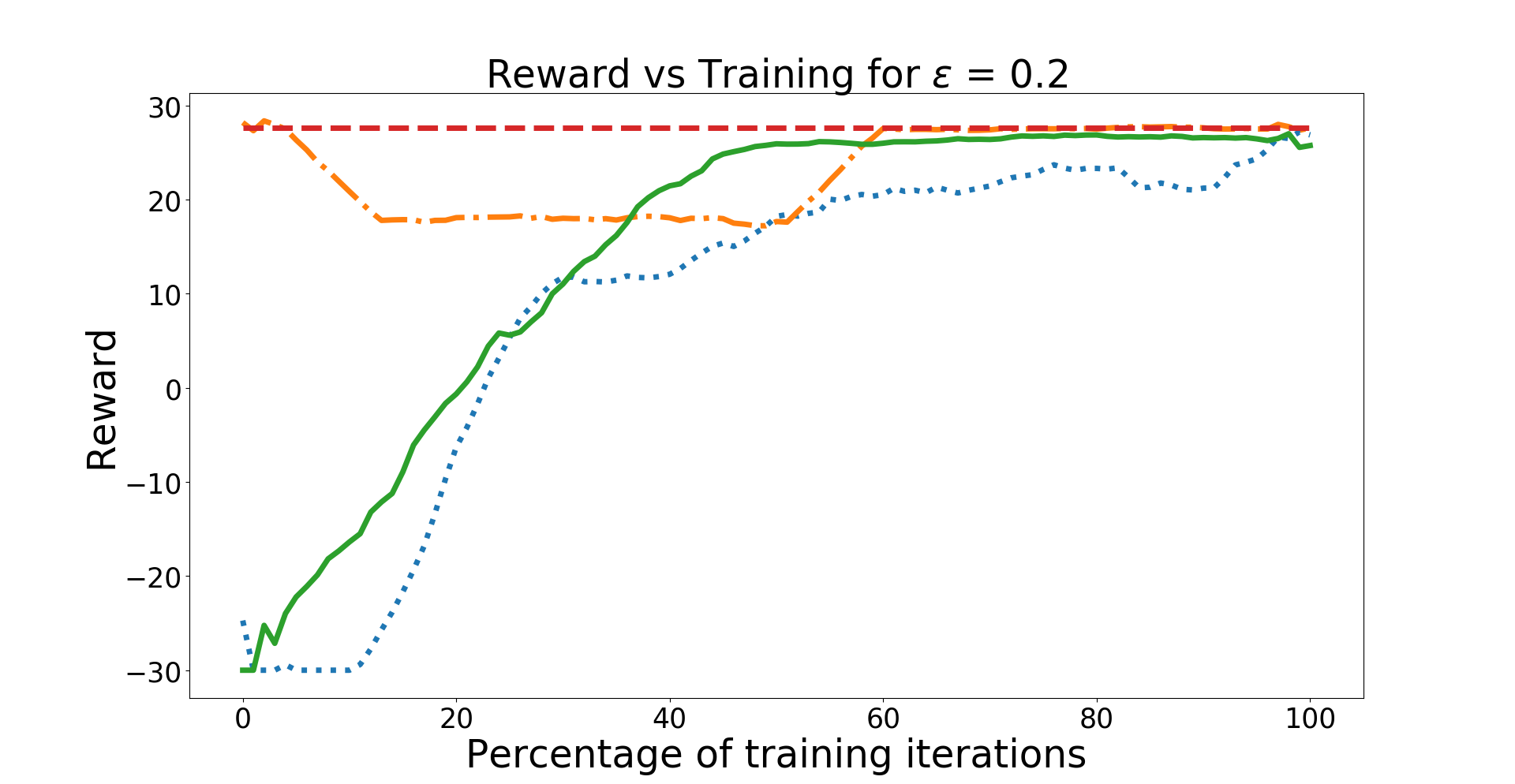}
                \phantomsubcaption
        \end{subfigure}
        
        \begin{subfigure}{0.32\textwidth}
                \includegraphics[width=\textwidth]{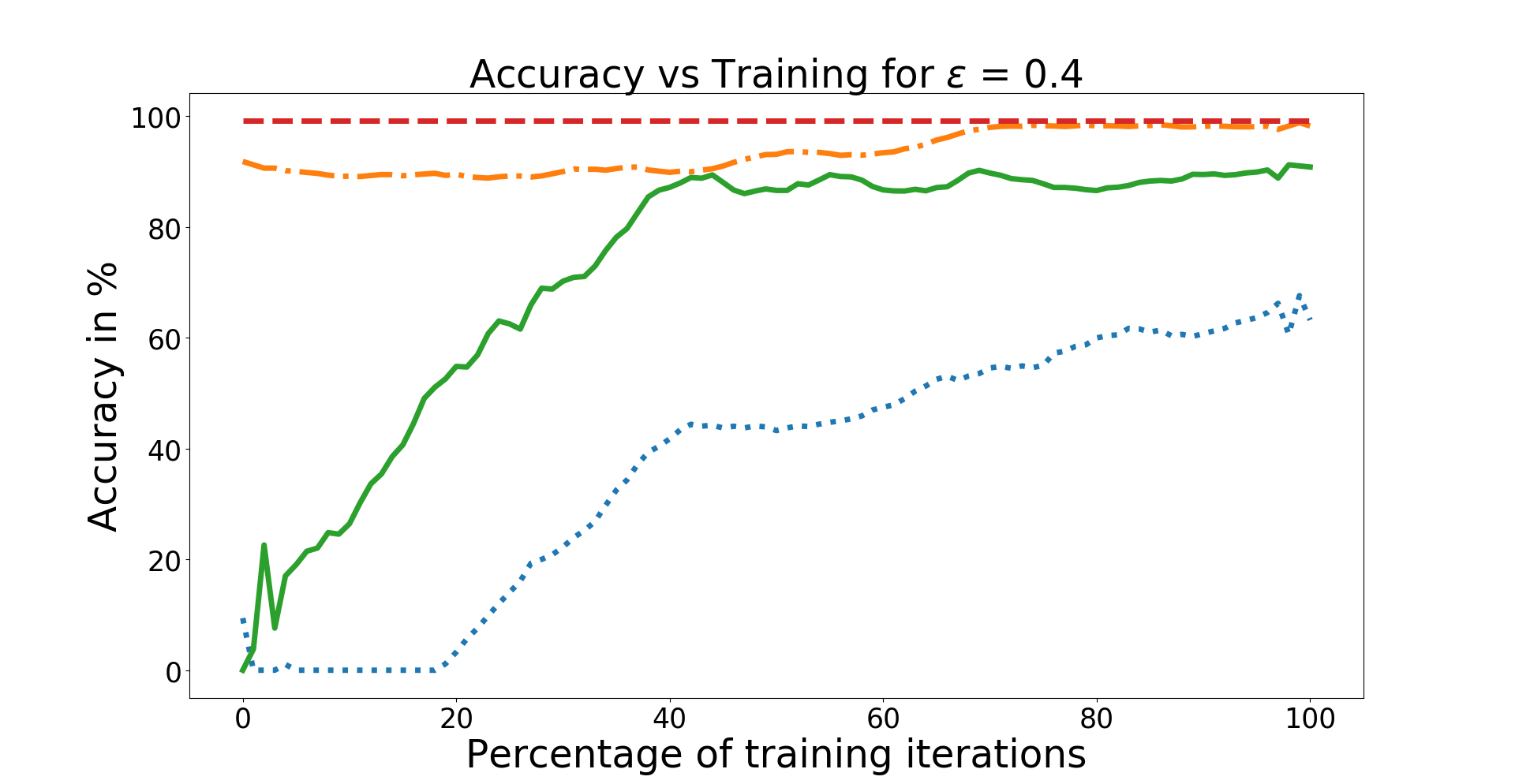}
                \phantomsubcaption
        \end{subfigure}
        ~
        \begin{subfigure}{0.32\textwidth}
                \includegraphics[width=\textwidth]{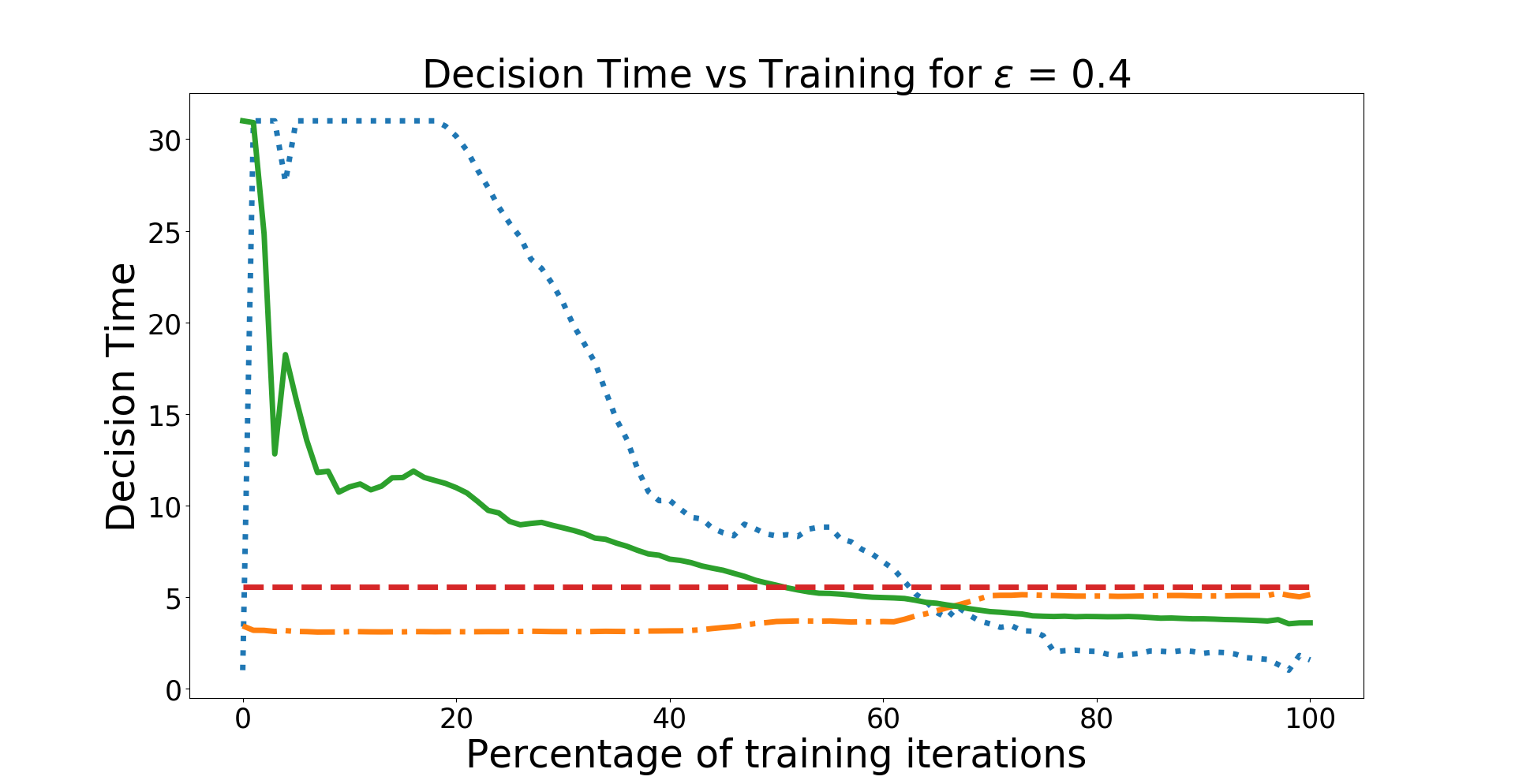}
                \phantomsubcaption
        \end{subfigure}
        ~
        \begin{subfigure}{0.32\textwidth}
                \includegraphics[width=\textwidth]{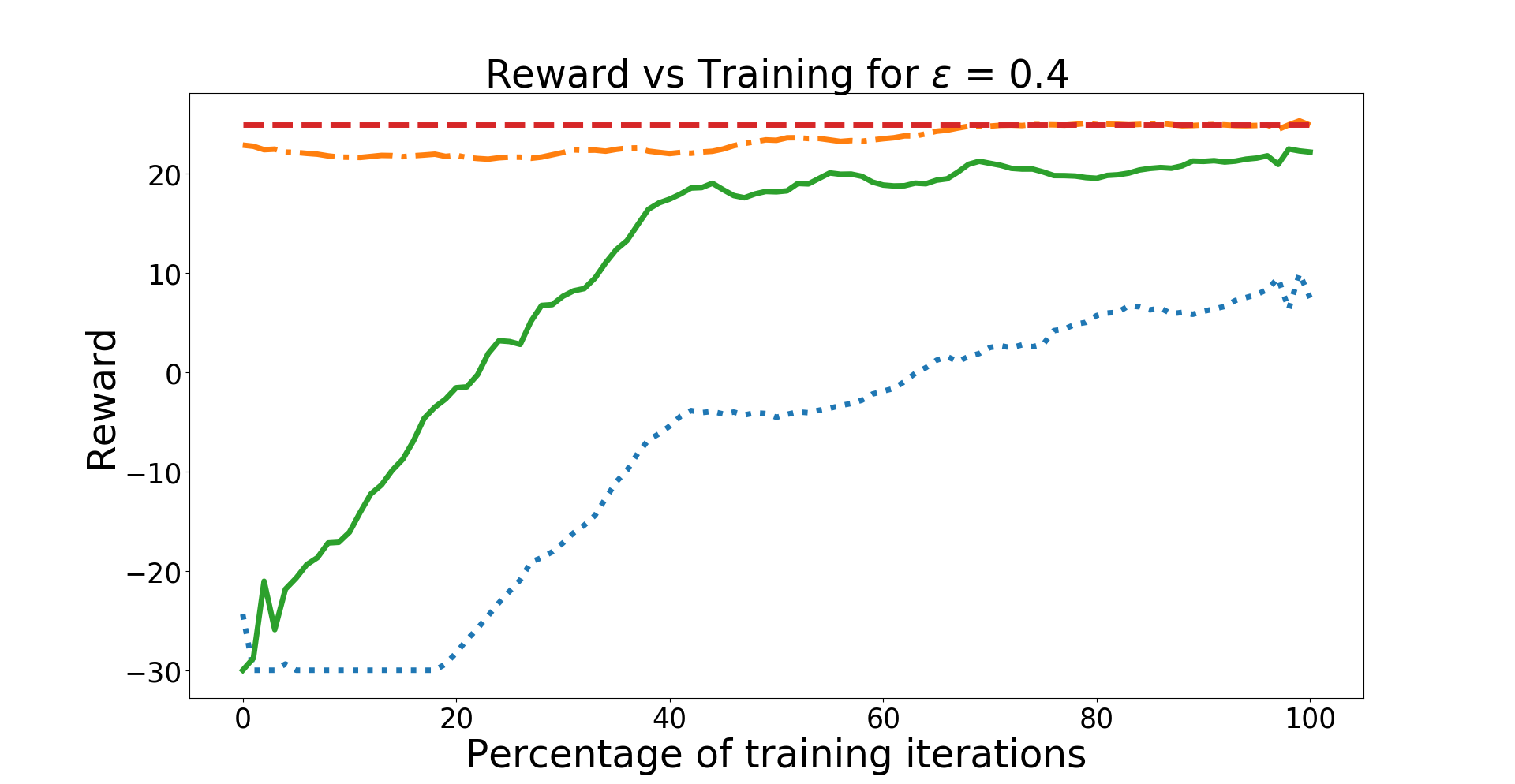}
                \phantomsubcaption
        \end{subfigure}
        
        \begin{subfigure}{0.32\textwidth}
                \includegraphics[width=\textwidth]{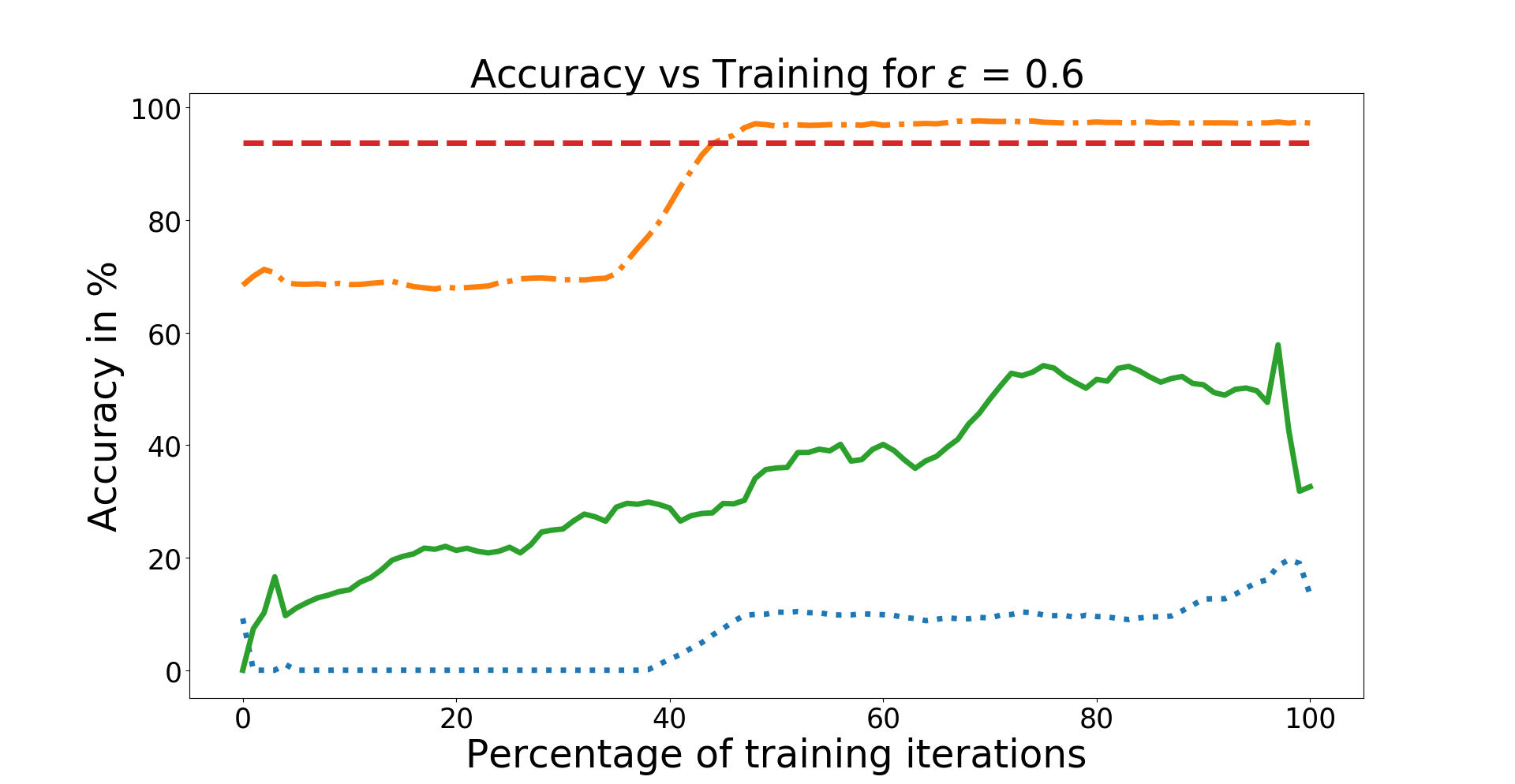}
                \phantomsubcaption
        \end{subfigure}
        ~
        \begin{subfigure}{0.32\textwidth}
                \includegraphics[width=\textwidth]{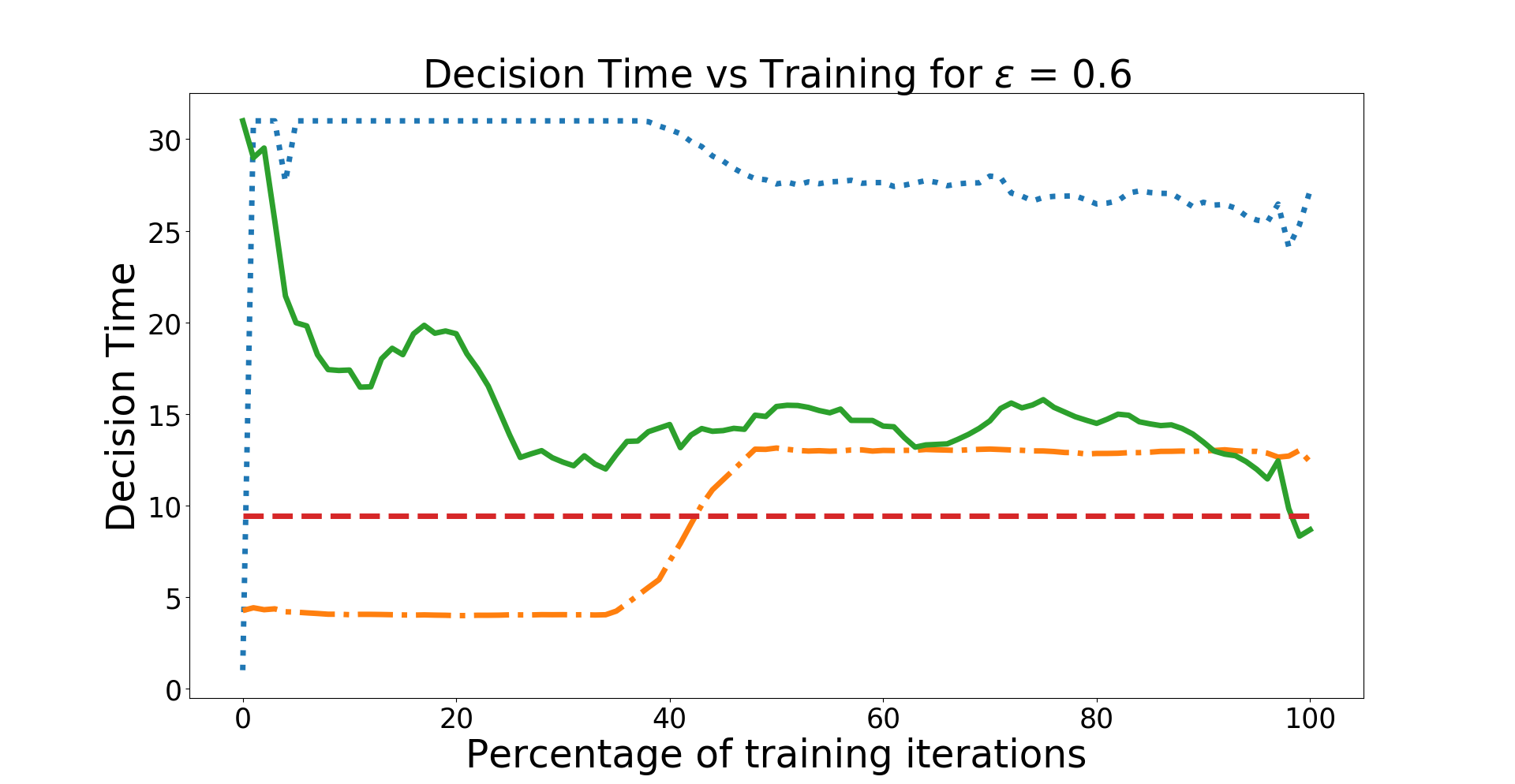}
                \phantomsubcaption
        \end{subfigure}
        ~
        \begin{subfigure}{0.32\textwidth}
                \includegraphics[width=\textwidth]{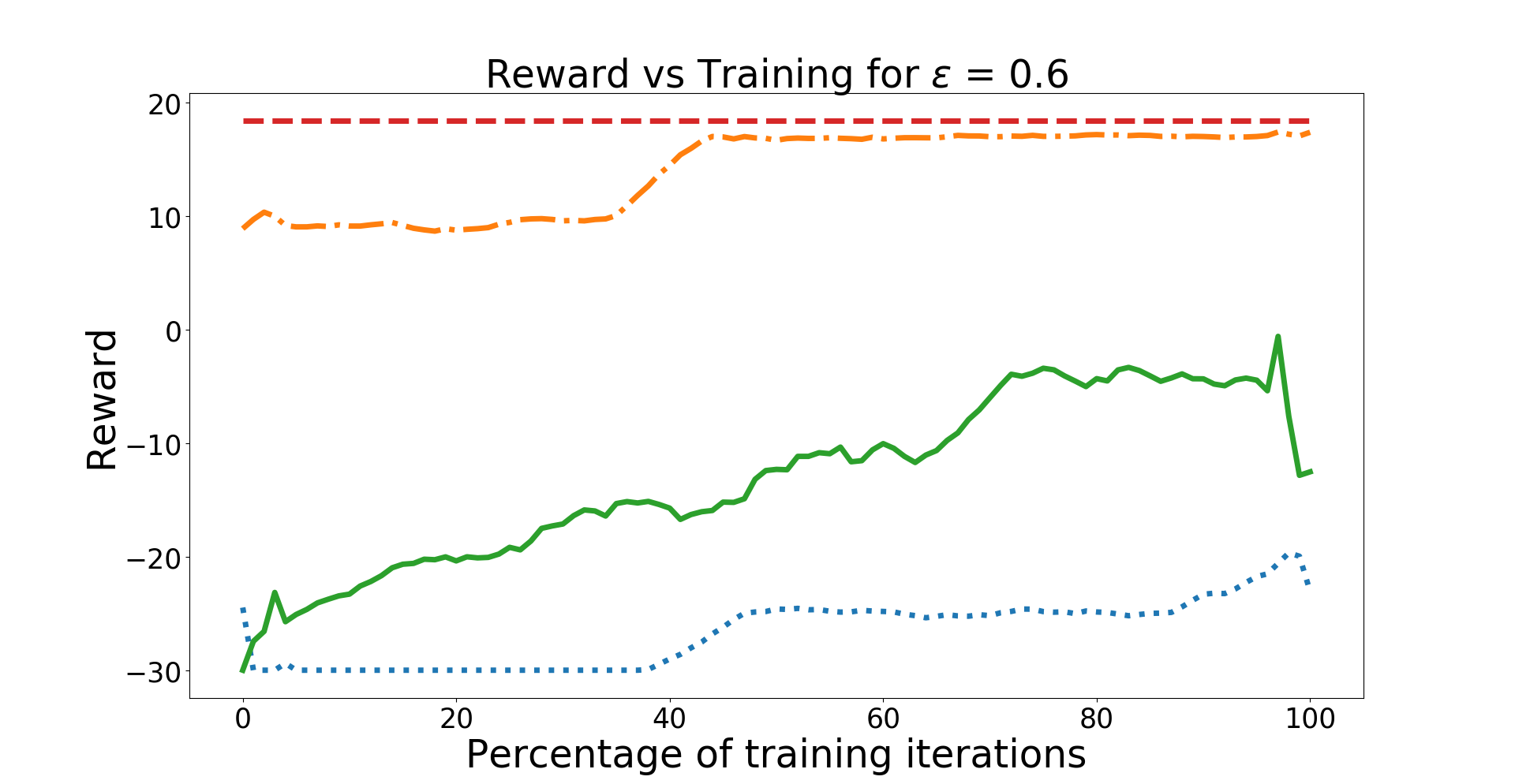}
                \phantomsubcaption
        \end{subfigure}
        
        \begin{subfigure}{0.32\textwidth}
                \includegraphics[width=\textwidth]{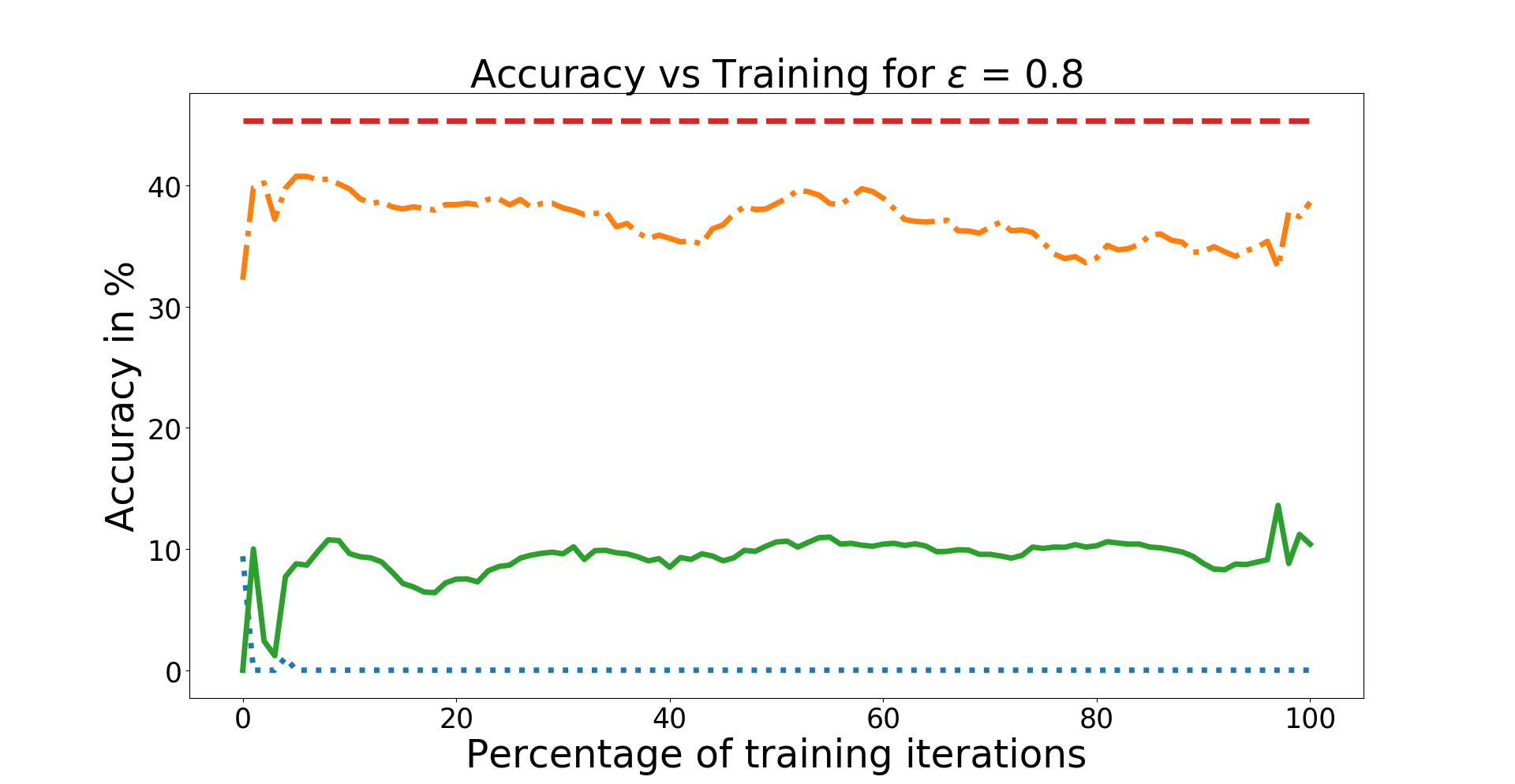}
                \phantomsubcaption
        \end{subfigure}
        ~
        \begin{subfigure}{0.32\textwidth}
                \includegraphics[width=\textwidth]{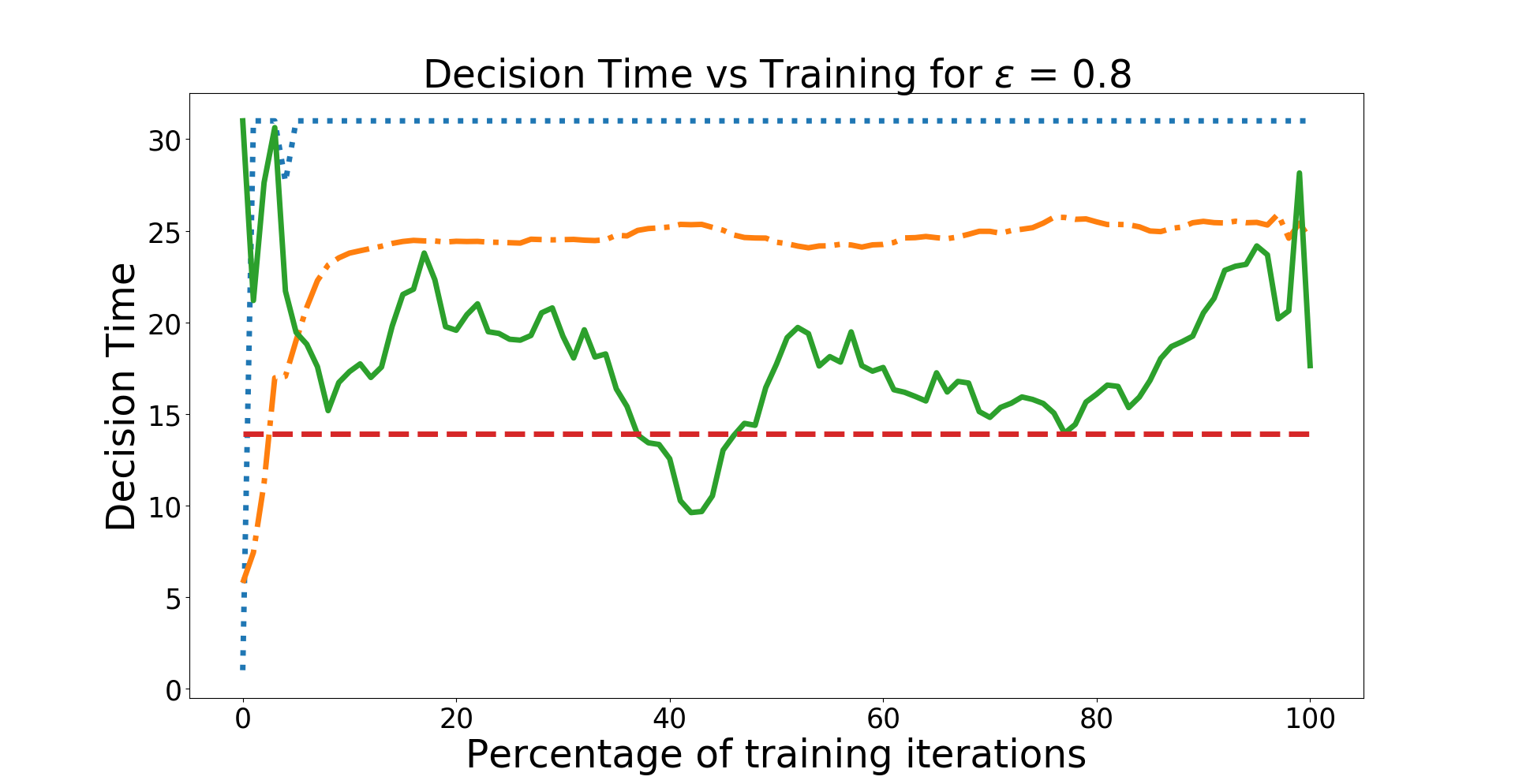}
                \phantomsubcaption
        \end{subfigure}
        ~
        \begin{subfigure}{0.32\textwidth}
                \includegraphics[width=\textwidth]{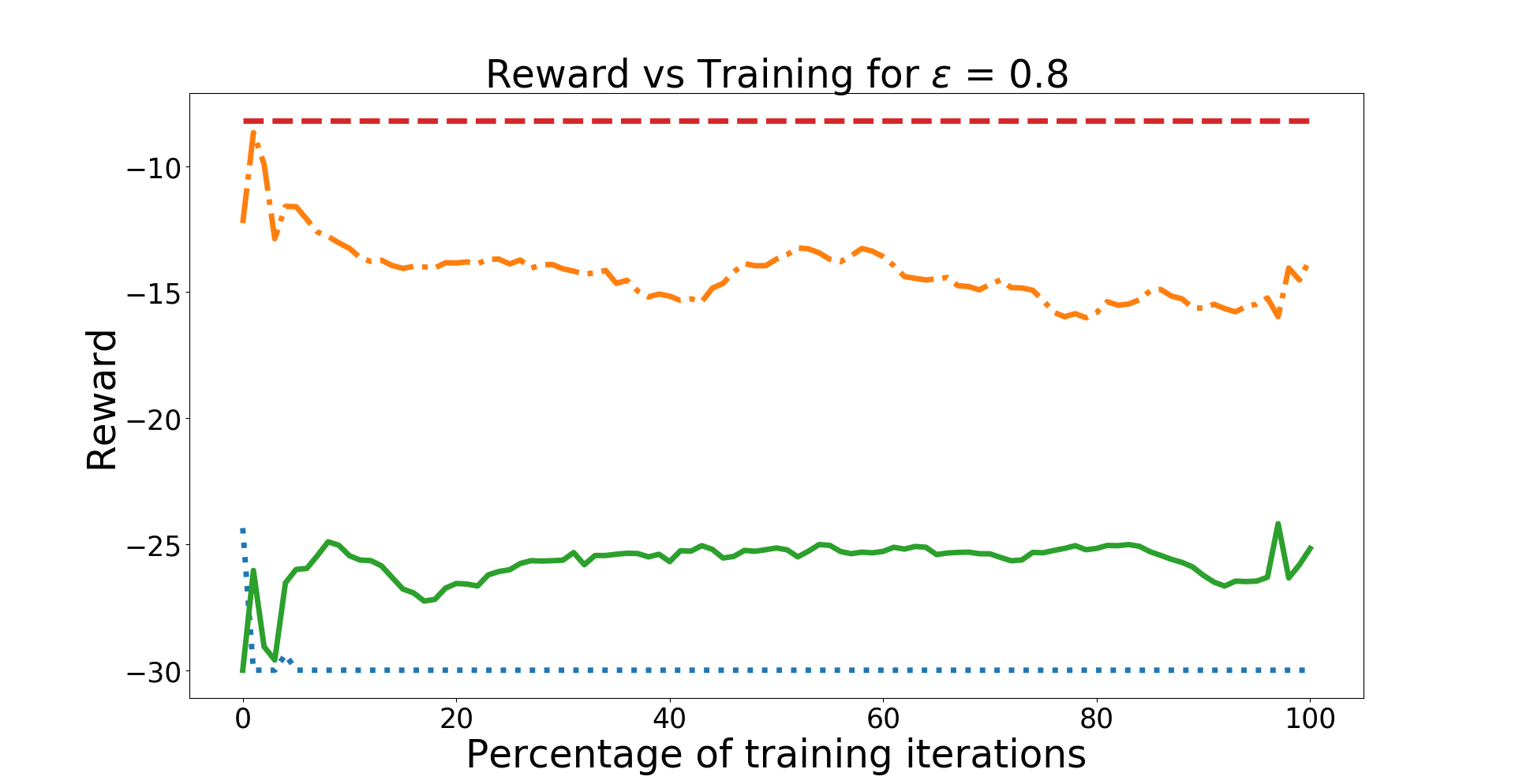}
                \phantomsubcaption
        \end{subfigure}
        \caption{Plots of Accuracy, Decision Time and Reward vs Training Iterations for 5 different values of $\epsilon \{0, 0.2, 0.4, 0.6, 0.8\}$. Each row has 3 plots for Accuracy, Decision Time and Reward (from left to right) for one value of $\epsilon$ (increasing from top to bottom). The plots for accuracy and reward often mirror each other but are both shown for completeness. As $\epsilon$ increases and the environment becomes more stochastic, the A2C-RNN agent heavily underperforms the accumulator module.}
        \label{fig:allresults}
\end{figure*}

The agent is trained for 50k episodes with entropy regularization with coefficient $\beta = 5.0$ to encourage exploration. Performance is evaluated after every 500 episodes.
The learning curves for expected accuracy, decision time and reward achieved by the learned network at each evaluation point are plotted in Fig. \ref{fig:allresults} using dotted blue lines.
Row 2 of Table \ref{tab:reward} presents results for the final expected rewards achieved by the policy trained with A2C-RNN. Using the Adam optimizer with learning rate $\num{1e-3}$, we find that while the A2C-RNN agent achieves near-optimal performance for low values of environment randomness ($\epsilon = \{0,0.2\}$), it's performance saturates at a lower reward level for $\epsilon = 0.4$ and it is unable to learn anything meaningful for $\epsilon = \{0.6,0.8\}$, with the expected reward not increasing from its initial value of -30, which is the lowest reward possible. This clearly shows that the A2C-RNN agent is unable to learn that it should wait, and make a safe decision only when it is confident in its chosen action. 
We hypothesize that this poor performance in the absence of the accumulator module, especially in environments with high uncertainty, is because learning to wait for long periods of time without having a built-in default `no-go' mechanism is difficult for any continuous parameterized function to learn, including an RNN. Intuitively, the agent would have to actively choose the 'No-Op' action for multiple time steps (say, the first 10 observations), and then, at the 11th observation, change it's neuron activations to choose the correct mode. In fact, the RNN would be required to have chosen `No-Op' when it received the exact same observation previously (but it's `confidence' was low).
This sudden change in the output, which has to be precipitated only by the cell state of the RNN (since the input form \textit{does not change}), is difficult for neural networks, which are continuous function approximators, to learn. 

\subsection{Learning Accumulator Threshold $\tau$}
\label{sec:tau}
Having established that a state of the art algorithm (A2C-RNN) with a forced action-selection policy is not able to do well at the Mode Estimation task, especially at high levels of environment randomness, we will show that replacing the traditional policy outputs of the actor-critic network with an accumulator module enables the agent to make safe decisions and achieve consistently high rewards in the ME task. In this section, we learn only the accumulator threshold $\tau$, verifying its utility - and in the next, use it to functionally replace the traditional outputs of a policy network.

The agent learns (using RL) a separate \textbf{Accumulator Network}, which predicts the optimal threshold as a function of the observation, $\tau = g(o)$. The observation $o$ received by the agent is a 10-dimensional one-hot vector representation of the sample, which is directly treated as the evidence to be accumulated (i.e., $\kappa = f(o) = o$). We note here that directly treating environment observations as evidence to be accumulated is possible only in simple environments as the one chosen here, but will not scale to more complex tasks, for which we jointly train both $f$ and $\tau$ (see Section \ref{sec:joint}). The approach used in this subsection, however, is presented as a minimal working example of the accumulator module.

The accumulator network's action space consists of 10 possible values for the threshold $\tau \in \{0.1,0.2,\ldots 0.9\}$, while the observations it takes in as input are a 10-dimensional one-hot vector representation of the sample. The observation is passed through a linear layer with a ReLU non-linearity to get an output of size 25, which is then passed as input to two linear layers that output the probability distribution over actions (choices of threshold) and the value estimate of the expected return. At every step, the agent's accumulated evidence is compared against the threshold decided by the accumulator network, and the agent makes a decision, or not, accordingly. 

The agent trains its accumulator network to choose a threshold that maximizes the rewards $r_t$ returned by the environment using the A2C algorithm and the Adam optimizer with learning rate $\num{1e-4}$, and includes entropy regularization with coefficient $\beta = 0.5$ to encourage exploration. It is trained for 50k episodes, with performance evaluated after every 500 episodes. The learning curves are plotted using dash-dot orange lines in Fig. \ref{fig:allresults}, while the final rewards are listed in Row 3 of Table \ref{tab:reward}. 
The agent is able to achieve optimal performance matching the MC simulations for all values of $\epsilon $ but $0.8$. It clearly outperforms the performance of the A2C-RNN agent, both in final performance and sample efficiency.

\subsection{Jointly Learning $\tau$ and Evidence Mapping $f$}
\label{sec:joint}
We have now established the viability of the proposed evidence accumulation mechanism; however, this instantiation comes with meaningful evidence (in the form of one-hot vector representations of the observation) received directly from the environment. In real situations, an agent will first need to extract evidence from its environment. For example, an agent receiving visual observations of its surroundings needs to extract evidence like objects, faces etc. from those images in order to present as evidence to the accumulator. As a simplified version of that we force the agent to learn a meaningful evidence mapping $f$ by providing it with 4-dimensional binary representations of the samples - requiring the agent to learn to extract evidence $\kappa$ for the 10 accumulator channels, while simultaneously learning the accumulator threshold $\tau$ - which is functionally analogous to replacing the traditional policy outputs in an actor-critic network with an accumulator module.

Here, we make two observations. First, the preference $\rho$ is calculated by a softmax over accumulator channels $\nu$, which means that $\rho$ is only a function of the differences between values in $\nu$. Hence, imposing a uniform lower bound on all the evidence vectors $\kappa_t$ does not restrict the performance of the accumulator. Second, since the highest threshold value we consider is 0.9, allowing arbitrarily large values in the accumulator channels $\nu$ is redundant since any evidence value $\kappa_t^i$ large enough such that $\rho^i$ exceeds 0.9 would be sufficient. Consequently, we are able to impose a loose upper bound on the evidence values $\kappa^i$, restricting $\kappa^i_t \in [0,S] ~\forall i,t$. We can further simplify this by restricting  $\kappa^i_t \in [0,1] ~\forall i,t$ and instead accumulating $\nu^i = \sum(S_t \kappa_t^i)$, where $1/S_t$ can be interpreted as the global sensitivity of the accumulator \textit{across channels}, and can be used to incorporate global action suppression mechanisms mirroring the hyper-direct pathway of the basal ganglia (section \ref{sec:cbgt}). For now, we treat $S_t$ as a hyperparameter that does not vary with time. 

We now sample each component of the evidence $\kappa_t^i \in [0,1]$ from a Beta distribution $\mathbb{B}_t^i$, whose concentration parameters $\alpha_t^i, \beta_t^i$ are predicted conditioned on the environment observation. The 4-dimensional (binary representations of samples) observations are passed through a linear layer of size 20, with ReLU non-linearity. The output is then passed as input to two linear layers that output the $\alpha$ and $\beta$ parameters, respectively, for all 10 components of the evidence vector, hence defining the distributions $\mathbb{B}_t^i$. We call this neural network, from which evidence $\kappa_t^i$ is sampled and accumulated in $\nu$, the \textbf{Evidence Network}.

Similar to the previous section, a separate \textbf{Accumulator Network} is responsible for deciding the accumulator threshold $\tau$, which follows the same architecture and training process as described there, except that we restrict the choice to 5 possible values of the threshold $\tau \in \{0.5,0.6,\ldots 0.9\}$. This ensures that there is always a single winner, since even with the lowest choice of $\tau=0.5$, no two components of the preference $\rho$ could exceed 0.5 at the same time.

After both the evidence $\kappa_t$ and threshold $\tau$ are obtained from the Evidence and Accumulator Networks, respectively, the agent accumulates evidence in its accumulator channels $\nu$, calculates the preference $\rho$, compares it with the threshold $\tau$ and accordingly decides whether or not to make a guess (act upon the environment). Both the evidence and accumulator networks are trained using the A2C algorithm, using the same reward (Eqn. \ref{eq:reward}). We use the Adam optimizer with learning rate of $\num{5e-4}$ for environments where $\epsilon = \{0,0.2\}$ and $\num{1e-3}$ when $\epsilon = \{0.4,0.6,0.8\}$. Entropy regularization is used for both networks, with coefficients $\beta$ 1.0 and 2.0, for the evidence and accumulator networks respectively. The agent trains for 50k episodes with evaluation every 500 episodes, and the learning curves have been plotted using solid green lines in Fig. \ref{fig:allresults}, while the final rewards are listed in Row 4 of Table \ref{tab:reward}. The jointly trained agent easily outperforms the A2C-RNN agent, learning greater patience and consequently winning greater reward. In the environments with $\epsilon = \{0.6,0.8\}$, where the A2C-RNN agent does not learn anything, the jointly trained agent learns even greater patience, and achieves significantly better performance than the A2C-RNN agent. 

\section{Conclusion}
% talking points: safe reinforcement learning, scaling up to full RL with state transitions, to more complex domains, discuss why RNNs don't perform well, talk about global suppression mechanisms to allow for context shifts and even more safety
In this paper, we propose a modification to existing RL architectures by replacing the policy/Q-value outputs with an accumulator module which sequentially accumulates evidence for each possible action at each time step, acting only when the evidence for one of those actions crosses a certain threshold. This ensures that when the environment is stochastic and uncertainty is high, the agent can exercise greater caution by postponing the decision to act until sufficient evidence has been accumulated, thereby avoiding catastrophic outcomes. We first define a partially observable task where the agent must estimate the mode of a probability distribution from which it is observing samples, and show that a state-of-the-art RL agent (A2C-RNN) is unable to learn even this simple task without an accumulator module, even though it is allowed to choose a `No-Op' action. We run Monte-Carlo simulations which provide baseline optimal estimates of the performance of the accumulator, and then learn the accumulator threshold as a function of the environment observations, showing that the accumulator module helps the agent achieve near-optimal performance on the task. Recognizing that in more complex real world tasks, the agent will have to extract meaningful evidence from the high-dimensional observations, we also jointly learn the evidence and the threshold, finding that this agent also easily outperforms the A2C-RNN agent, while being equally or more sample efficient. 

These results make a strong case for adding an accumulator module to existing Deep RL architectures, especially in real-world scenarios where individual observations are incomplete and unreliable, the cost of making a bad decision is very high, and longer decision times are an acceptable price to pay for assurances that those decisions will be both safe and accurate. 

\section{Future Work}
The Mode Estimation task as defined in this paper is, essentially, a partially observable multi-armed contextual bandit. While the context (the mode of the distribution) is unknown to the agent, it does not transition to different contexts within an episode, as is common in reinforcement learning tasks. We plan to test the accumulator module on tasks with state transitions, and then on more complex domains (such as the Atari games \cite{bellemare2013arcade}). Another interesting line of work is to add a global suppression mechanism (similar to the hyperdirect pathway in the CBGT, see Section \ref{sec:cbgt}), by allowing the agent to change the sensitivity across accumulator channels based on environmental signals. Having a global stopping mechanism would be very useful for agents operating in very dynamic and reactive environments, such as self-driving vehicles on open roads.

\subsubsection*{Acknowledgments}
This research was sponsored by AFOSR Grants FA9550-15-1-0442 and FA9550-18-1-0251. 
\bibliography{refs}
\bibliographystyle{aaai}

\end{document}